\documentclass[lettersize,journal]{IEEEtran}
\usepackage{amsmath,amsfonts}
\usepackage{algorithmic}
\usepackage{algorithm}
\usepackage{array}
\usepackage[caption=false,font=normalsize,labelfont=sf,textfont=sf]{subfig}
\usepackage{textcomp}
\usepackage{stfloats}
\usepackage{url}
\usepackage{verbatim}
\usepackage{graphicx}
\usepackage{cite}
\usepackage{booktabs}
\usepackage{multirow}
\usepackage{color}
\begin{document}

\title{Exposing AI-generated Videos: A Benchmark Dataset and a Local-and-Global Temporal Defect Based Detection Method}

%\author{IEEE Publication Technology,~\IEEEmembership{Staff,~IEEE,}
        % <-this % stops a space
%\thanks{This paper was produced by the IEEE Publication Technology Group. They are in Piscataway, NJ.}% <-this % stops a space
%\thanks{Manuscript received April 19, 2021; revised August 16, 2021.}}
\author{Peisong He, Leyao Zhu, Jiaxing Li, Shiqi Wang,~\IEEEmembership{Senior Member,~IEEE}, Haoliang Li,~\IEEEmembership{Member,~IEEE}
\thanks{This work is supported by National Key Research and Development Program for Young Scientists under Grant 2022YFB4501300. (Corresponding author: Dr. Haoliang Li.)}
\thanks{P. He and L. Zhu are with School of Cyber Science and Engineering, Sichuan University, Chengdu, China (e-mail: gokeyhps@scu.edu.cn, {zhulyscu@gmail.com}). }\thanks{J. Li is with the Department of Electrical and Computer Engineering, Carnegie Mellon University, Pittsburgh, US (email: jiaxingl@andrew.cmu.edu).}\thanks{S. Wang is with the Department of Computer Science, City University of Hong Kong, Kowloon, Hong Kong (e-mail: shiqwang@cityu.edu.hk).}\thanks{H. Li is with the Department of Electrical Engineering, City University of Hong Kong, Hong Kong (e-mail: haoliang.li@cityu.edu.hk).}}

% The paper headers
%\markboth{Journal of \LaTeX\ Class Files,~Vol.~14, No.~8, August~2021}%
%{Shell \MakeLowercase{\textit{et al.}}: A Benchmark for AI-generated Videos}

%\IEEEpubid{0000--0000/00\$00.00~\copyright~2021 IEEE}
% Remember, if you use this you must call \IEEEpubidadjcol in the second
% column for its text to clear the IEEEpubid mark.

\maketitle

\begin{abstract}
%The generative model has made significant advancements in the creation of realistic images and has also demonstrated its efficacy in generating videos. Consequently, there is a growing need for detection systems capable of distinguishing between AI-generated videos and natural ones to mitigate the spread of misinformation. However, this emerging risk has not been adequately addressed due to the absence of a comprehensive dataset containing a diverse range of videos generated by various advanced algorithms. This paper introduces the Imposter-Video dataset, which serves as a benchmark for video analysis. The dataset is notable for its: 1) ample size, comprising a total of 5000 videos suitable for training and testing 3D Resnet models; 2) diverse range of scenarios, encompassing 1000 scenarios across various contexts; 3) inclusion of outputs from four distinct, widely used generators; and 4) examination of the effects of post-processing induced by propagation. The Imposter-Video dataset serves as a valuable resource for researchers seeking to further investigate the detection of AI-generated videos.
The generative model has made significant advancements in the creation of realistic videos, which causes security issues. However, this emerging risk has not been adequately addressed due to the absence of a benchmark dataset for AI-generated videos. In this paper, we first construct a video dataset using advanced diffusion-based video generation algorithms with various semantic contents. Besides, typical video lossy operations over network transmission are adopted to generate degraded samples. Then, by analyzing local and global temporal defects of current AI-generated videos, a novel detection framework by adaptively learning local motion information and global appearance variation is constructed to expose fake videos. Finally, experiments are conducted to evaluate the generalization and robustness of different spatial and temporal domain detection methods, where the results can serve as the baseline and demonstrate the research challenge for future studies.
\end{abstract}


\section{Introduction}
%\IEEEPARstart{T}{he} progression of video synthesis has evolved from producing basic animated digital characters to generating photorealistic videos. The emergence of text-to-video models has notably decreased the complexity associated with creating surreal videos, in contrast to previous deep fake techniques, as it enables users to generate a diverse array of scenarios rather than being limited to specific settings. Despite the potential concerns associated with this advancement, the current lack of an effective detection system to address this issue can primarily be attributed to the absence of a suitable benchmark dataset. 

Recently, the explosive growth of artificial intelligence generated content (AIGC) has catalyzed revolutionary developments in several fields, such as social media and entertainment industry. Due to its impressive controllability and diversity \cite{li2023aigc}, researchers are starting to pay more attention to diffusion model for the generation of images \cite{chen2024t2i,chen2024restortion} and videos\cite{khachatryan2023text2video}.
%than generative adversarial networks \cite{KarrasALL18,BrockDS19} and auto-regressive Transformers \cite{lee2022autoregressive}. Diffusion models are a type of probabilistic generative models, which include a diffusion process gradually adding noises and an inverse process to denoise the degraded data to restore the input image. 
Compared with static images, videos contain temporal dynamic information to present better visual experience. Video diffusion models \cite{HoSGC0F22} are progressively used on several video-related applications, including video generation, which has evolved from producing basic animated digital characters to generating photorealistic videos. Among these applications, text-to-video generation garners significant attention, which can automatically generate videos controlled by text prompts. 
%It has notably decreased the complexity of creating realistic videos, which enables users to generate a diverse array of scenarios rather than being limited to specific settings \cite{liu2023deepfacelab}. 

On the other hand, cybercriminals can use video generation technologies to produce more realistic fake videos and online platforms, such as social networks, simplify the broadcasting and sharing of videos, posing severe negative impacts on public safety.
%On the other hand, using video generation technologies by cybercriminals enables the production of more realistic fake news, posing severe negative impacts on public safety.
Therefore, there is an urgent need to study corresponding countermeasures. In contrast to video editing software, text-to-video generation methods are capable of modeling both spatial and temporal long-range relationships. Consequently, existing video forensic datasets are insufficient to satisfy the demands for forensic analysis of AIGC videos created by text-to-video generators and there is a lack of forensics tools designed to identify AI-generated videos.

%In this study, we have curated a collection of widely used video diffusion generators sourced from Hugging Face, as well as the text2video zero[] model, which utilizes a pre-trained image diffusion model for video generation, in order to construct our AI-generated surreal video dataset, referred to as Imposter Video. Subsequently, utilizing our dataset as a foundation, we have employed state-of-the-art video classification models to conduct two sets of domain-cross experiments aimed at examining the generalization capabilities of the classifiers. The first set entails Generator Cross experiments, involving the training and testing of the classifier on groups of videos generated by different generators. The second set involves Post Processing Cross experiments, which simulate the detector processing videos degraded by propagation. To achieve this, we implemented three processing methods from the Corruption Robustness Benchmark[], specifically H.265 ABR Compression, H.265 CRF Compression, and Bit Error, to degrade the quality of the test videos.

To address the issues above, we first construct an AI-generated video dataset, which considers several advanced AI video generation algorithms to create fake videos with various contents. Besides, a novel detection framework based on local and global temporal defects is proposed. Based on this dataset, cross-domain evaluation are considered to analyze generalization and robustness of different methods. The main contributions of this paper are summarized as follows:
\begin{enumerate}
\item[1)] We propose an AIGC video forensic dataset incorporating a variety of video diffusion generators and zero-shot text-to-video generators. Various text prompts are used to ensure the diversity of spatial and temporal contents of generated videos. Besides, typical video lossy operations are considered to create degraded samples to simulate network transmission process.
\item[2)] According to the analysis of temporal defects in generated videos, we proposed a novel detection framework by considering local motion information and global appearance variation. Besides, channel attention based feature fusion is designed to expose fake video by combining local and global temporal clues adaptively.
%We propose the XXX detection algorithm, which is compared with several state-of-the-art video classification networks as baseline methods. Building upon the experimental results, we also outline the challenges and future research directions in AIGC video forensics.

\item[3)] Extensive experiments are conducted to evaluate the generalization capability in cross-generator scenarios and the robustness against lossy operations, including video compression and transmission error. Experimental results can be used as the baseline for future work.

%Based on the diversity of generation methods, we design a cross-generator detection protocol, which can be used to evaluate the generalization capability of detection methods. Common lossy operations in video transmission are considered, including H.265 ABR compression, H.265 CRF compression, and bit error, which aims to simulate the degradation caused by video transmission to evaluate the robustness of detectors.
\end{enumerate}

\section{Related Works}
%(can be expended into individual sections)
%Text-to-image generator models like stable diffusion and mid-journey have achieved generating photorealistic images, their success also attracted people's concern about the spread of misinformation, and since then various datasets like GenImage[1] have been published to support the development of detection algorithm, and a lot of successful detection algorithms like Spec[2]. And the image generators’ success has been migrated to video generation through various methods. MAGVIT[3] leveraged the transformer’s advantage of learning temporal correlation in natural language and applied it to learn the correlation between different frames through a 3D tokenizer， while the video diffusion[4] expanded the Image Diffusion model to a 3D U-net architecture, both of the previous 2 approaches require an enormous amount of training data Text2Video-Zero[5] tried to reduce the required training computation by leveraging the pretrained image diffusion model with a devised temporal continuation algorithm. Although the video generators are achieving a more realistic result, there is no appropriate benchmark dataset for researchers to develop efficient detectors, and our work intends to mitigate that gap. 
\subsection{Video Generation Algorithm}
The diffusion model was originally proposed for image generation \cite{ho2020denoising}, which includes a forward diffusion process and a backward denoising process. Diffusion-based image generation models can be controlled by different types of information, such as text prompts. Since video data can be treated as a sequence of images, Ho et al. proposed the earliest diffusion-based video generation algorithm, video diffusion model \cite{HoSGC0F22}, by extending the 2D U-Net structure of the traditional image diffusion models to a 3D structure. It can capture the spatial and temporal features jointly from video data. Then, Zhou et al. \cite{zhou2022magicvideo} constructed the prior work, MagicVideo, to leverage latent diffusion model (LDM) for text-to-video generation in the latent space, which improves the generation efficiency. \textcolor{black}{More recently, the researchers \cite{luo2023videofusion,alivilab} incorporated spatial-temporal convolution and attention into LDMs for text-to-video tasks.} Khachatryan et al.\cite{khachatryan2023text2video} utilized the pre-trained stable diffusion model to conduct zero-shot video generation, which reduces computational costs.
%performed a cross-attention mechanism to ensure consistency between frames and reinforced latent code with motion dynamics. 
%Huang et al. \cite{huang2023freebloom} introduced large language models to generate frame-level prompts to guide LDM to produce coherent frames. 
%Their proposed Free-Bloom utilized joint noise sampling and step-aware attention shifting to improve temporal and identical coherence while maintaining semantic coherence.

\subsection{AIGC Multimedia Forensics Dataset}

As generative networks became capable of producing high-resolution realistic images, forensic analyzers began to pay attention to the detection of AI-generated images \cite{chen2022csvt,WangW0OE20,2312-10461}. In \cite{WangW0OE20} and \cite{ZhuCYHLLT0H023}, authors constructed large-scale AI-generated image datasets, including GAN-based and diffusion-based generators. However, unlike static images, video data contains temporal information, that provides dynamic visuals. Besides, videos are stored and broadcast using lossy compression techniques, which reduce temporal redundancy by the inter-frame coding process, such as motion compensation. Therefore, there is an urgent need in the field of video forensics to construct an AI-generated video dataset.

\section{AI-generated Video Dataset}
\subsection{Data Collection}
The natural videos were collected from the Microsoft Research Video Description Corpus (MSVD) dataset \cite{chen2011collecting}, where videos in the MSVD dataset were crawled from YouTube. The MSVD dataset was originally constructed to study video captioning techniques, and each video has a description. Therefore, these associated captions of videos can be utilized as text prompts for the generation of fake videos (negative samples) by video generators. Specifically, the dataset comprises 1000 natural samples paired with 1000 negative samples for each generator, {where each negative sample consists of 24 frames. In this dataset, four advanced diffusion-based video generators are considered, which obtains 96k fake frames in total.}
\begin{figure}[!t]
\centering
\includegraphics[width=3.5in]{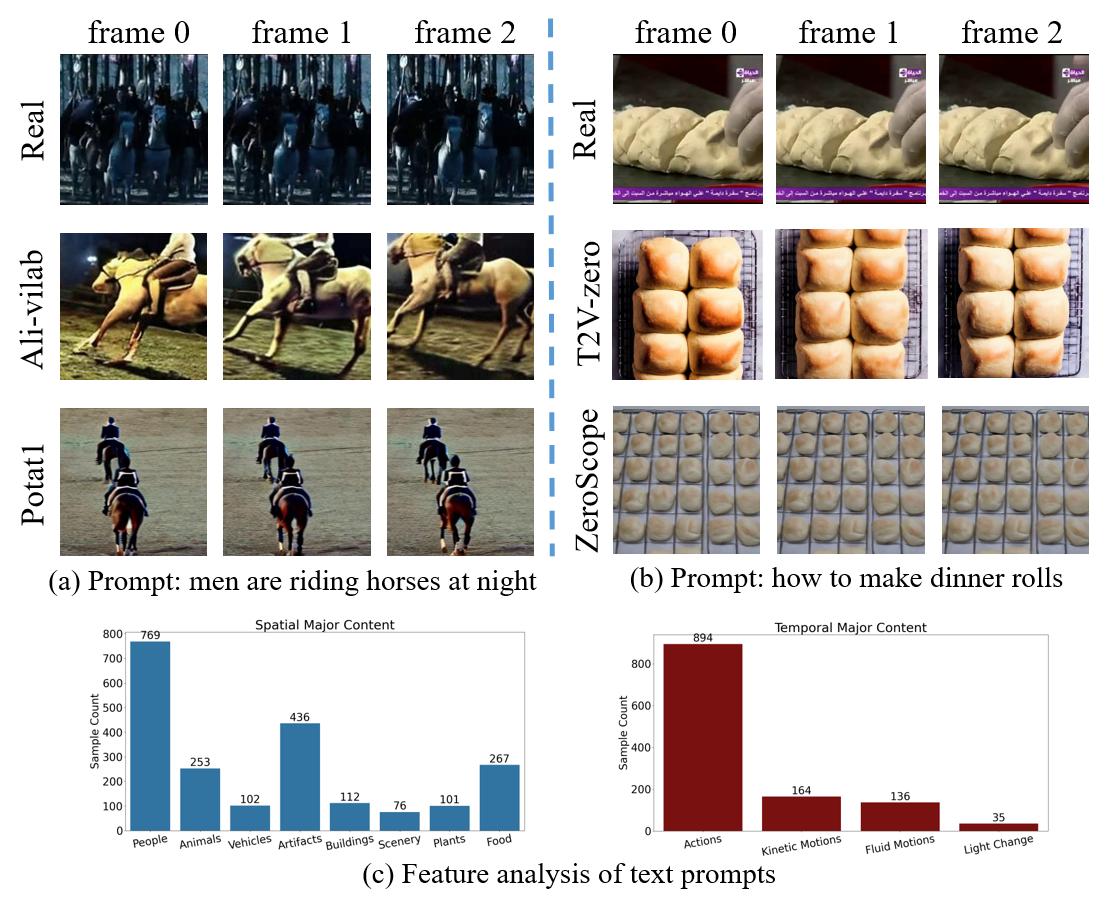}
\caption{Examples of AI-generated videos and feature analysis of our dataset.}
\label{fig:example}
\end{figure}

\subsection{Feature Analysis}
%The proposed AI-generated video dataset is organized and analyzed. Its distinct features are summarized as follows:
The distinct features of the proposed dataset are analyzed.
\subsubsection{Various Contents} We first analyze the semantic diversity for text prompts, which are used to control the contents of generated videos. Similar to \cite{NEURIPS2023_c481049f}, the statistical analysis is conducted in aspects of spatial and temporal contents, respectively. As shown in Fig. \ref{fig:example}, spatial major contents can be summarized into eight groups, such as people and buildings. For temporal content, generated videos have rich motion information, which differ from static images.

\subsubsection{Various Video Generators}

In this dataset, various generators based on video diffusion model \cite{HoSGC0F22} and a zero-shot text-to-video model are considered. Specifically, three most popular and publicly released video diffusion models are selected, including \textcolor{black}{ali-vilab} \cite{luo2023videofusion,alivilab}, zeroscope \cite{zeroscope}, and potat1 \cite{potat1}. Among them, \textcolor{black}{ali-vilab} is a text-to-video generation model using spatio-temporal blocks to conduct frame generation with temporal consistency and smooth motions. Following the similar generation pipeline, zeroscope and potat1 considers more fine-tuning strategies and other generation configurations, such as spatial resolutions. Besides, a zero-shot text-to-video generation method\cite{khachatryan2023text2video} is also considered, which utilizes the pre-trained image diffusion model and keeps temporal consistency with motion dynamics.
%to improve computational efficiency and mitigate data hungry problem.
\subsubsection{Video Lossy Operations} To simulate lossy operations during network transmission \cite{Wells2017Error,Li2023live}, three kinds of video post-processing operations are considered to generate videos with quality degradation for robustness evaluation, including H.265 Average Bit Rate (ABR) compression, H.265 Constant Rate Factor (CRF) compression, and Bit Error. Each operation applies various degrees of operation intensity. Detailed settings are presented in Section \ref{sec:lossy}.

%Each generator has been assigned a schedule consisting of 25 inference steps. An inherent limitation of the video diffusion model is its demanding training process, which necessitates a substantial amount of data and computational power. 

%In summary, four AI-generated methods are considered in this paper. All these generators are given the same set of 1000 text prompts to generate fake videos.
.

\section{Detection method}
\begin{figure*}[!t]
\centering
\includegraphics[width=0.9\textwidth]{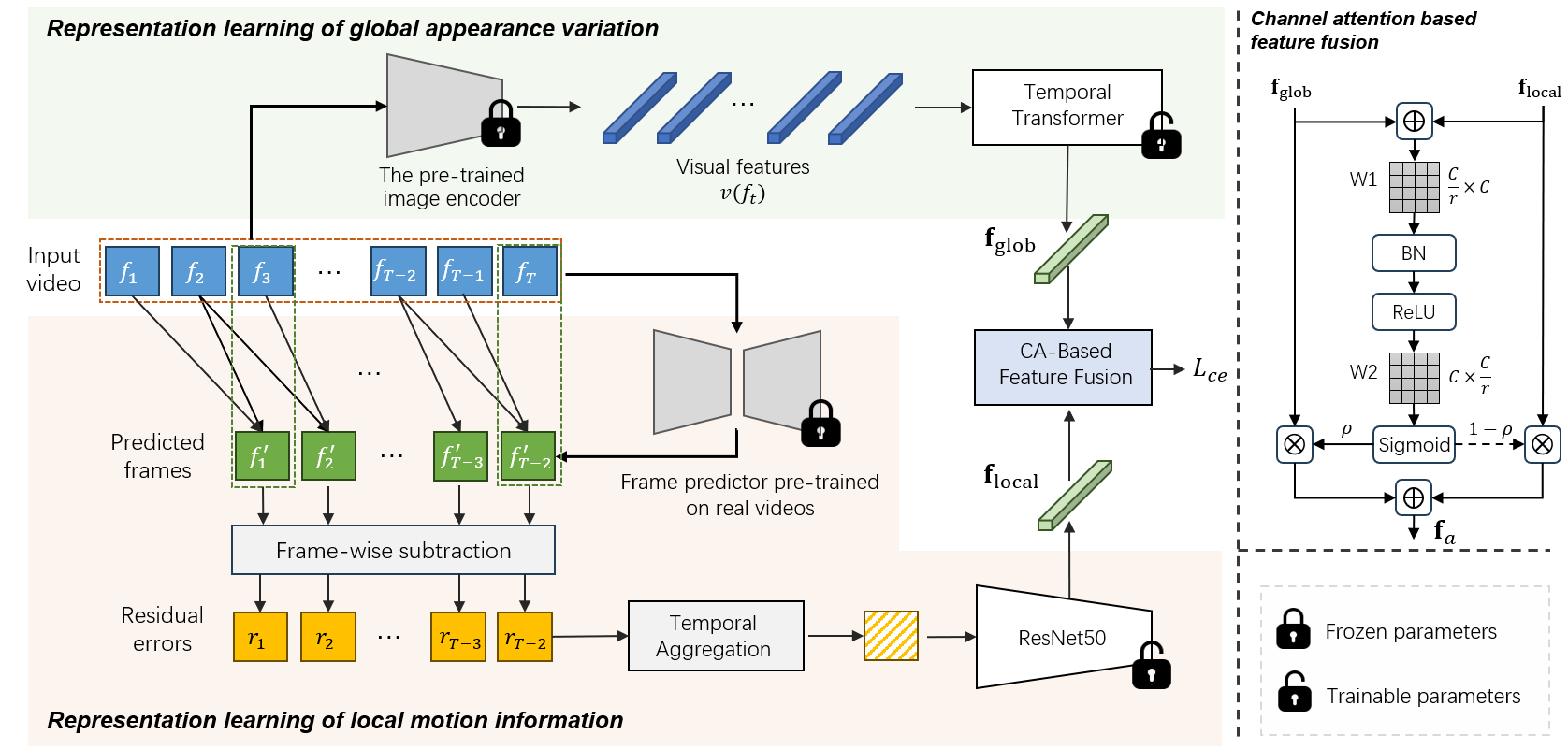}
\caption{The proposed AI-generated video detection framework based on local and global temporal defects.}
\label{fig:framework}
\end{figure*}

%\subsection{Overview}
AI-generated videos show different temporal dependencies compared with real videos, since their capturing and generation processes are dissimilar. Specifically, real videos are captured by camera devices, where the short time interval between adjacent frames leads to a very high temporal redundancy.
%the time resolution of a real videos is determined by the frame rate of the camera. The frame rate adopted by mainstream video collection devices is 25fps or above [xx], which means the time interval between temporally adjacent frames is extremely short (e.g., 40ms) and results in very high temporal redundancy. 
On the other hand, AI video generators control the time continuity of frames in the latent space. %For example, ModelScope designs a U-Net equipped with spatio-temporal blocks to capture motion continuity in the latent space. 
These differences result in the defects of generated videos at different spatio-temporal scales. In this section, the proposed AI-generated video detection framework based on local and global temporal defects is presented in detail, as shown in Fig. \ref{fig:framework}.

%The first branch concentrates on motion learning, while the other branch focuses on appearance variation learning.

\subsection{Representation Learning of Local Motion Information\label{sec:local}}
Due to the limitation of algorithm and computational cost, existing AI video generators are still hard to fully model the temporal characteristics of real videos. We tackle the problem of exposing abnormal motion patterns in AI-generated videos by measuring the predictability of inter-frame information.
%, where temporal dependency of adjacent frames are considered. 
%Inspired by the above analysis, we leverage the predictability of inter-frame motion as a generalized clue for the detection of AI-generated videos in the temporal domain. 
%In the branch of learning local motion information, we propose a temporal ``prediction-aggregation’’ network, which first measures temporal dependency using the predictability of inter-frame motion, and then aggregates prediction errors along the temporal dimension to obtain long-range information and mitigate the overfitting problem of spatial details.

\subsubsection{Extraction of Frame Prediction Error\label{sec:fpe}}
%For real videos, the high sampling rate in the time domain leads to the strong temporal redundancy between adjacent frames, which provides the foundation for inter-frame prediction. 
Inspired by the idea from anomaly detection \cite{NayakPD21}, a frame predictor $P(\cdot,\cdot)$ is first trained by only using real videos to learn their ``normal'' motion patterns in local regions. Then, the parameters of this frame predictor are frozen. For an input video $\mathbf{F}=[f_1, f_2,\dots, f_T]$, two adjacent frames $(f_t, f_{t+1})$ are fed into $P(\cdot,\cdot)$ to obtain the predicted frame $\tilde{f}_{t+2}$. Prediction errors can be calculated by $r_{t}=\tilde{f}_{t}-f_{t}$ and the sequence of prediction errors $\mathbf{R}$ is obtained ($\mathbf{R}=[r_1,r_2,...,r_{T-2}]$), which are used to measure the predictability of inter-frame local motion. 

\subsubsection{Temporal Aggregation}
For real videos, the amplitude of their prediction errors is relatively small compared with fake videos due to high temporal redundancy. However, rich and diverse spatio-temporal contents may degrade the generalization capability of learned features. To address this issue, a temporal aggregation operator $A(\cdot)$ is designed for the sequence of prediction errors: $A(\mathbf{R})=\hat{R}$ and $\hat{R}(i,j)=\frac{1}{T-2}\sum_{t=1}^{T-2} r_{t}(i,j)$. This simple and efficient strategy is applied to maintain long-range information and also mitigate the influence of various spatio-temporal details. Then, this aggregated prediction error map $\hat{R}$ is fed into a 2D encoder ($F_l(\cdot)$), such as ResNet50, to obtain the local motion feature, $\mathbf{f}_{local}=F_l(\hat{R})$.

\subsection{Representation Learning of Global Appearance Variation\label{sec:global}}
%Except for abnormal local motions, As discussed in Section xx, different from the temporal sampling process of video capture devices, most AI video generators model the temporal continuity in the latent space. Due to the 

Although AI video generators can model temporal continuity in the latent space, the lack of strong constraints for inter-frame consistency may cause abnormal variations of object appearances. For example, in the second row of subfigure \ref{fig:example}(a), there is a distinct variation of person's waist-to-leg proportions between ``frame 0'' and ``frame 1''. We call this type of defects as abnormal global appearance variation. 

To explore this clue, visual features of input frames are first extracted by a pre-trained image encoder, and then a transformer is applied to model their temporal variations.
%, where the frame-wise visual features are treated as tokens. 
It should be noted that we adopt the pre-trained image encoder on vision tasks instead of training it directly on fake frames, which aims to mitigate the overfitting issue to specific generation patterns. Specifically, BEiT v2 \cite{2208-06366} is used as the pre-trained visual feature extractor in this work, which designs a masked image modeling framework. Its vector-quantized knowledge distillation can reduce the sensitivity of visual features to changes in image details and present high-level semantics.

%BEIT was initially proposed to solve the data hungry problem of visual transformers (ViTs), which designs a masked image modeling framework in a self-supervised learning manner. It first constructs a vision tokenizer to convert each image to a set of discrete vision tokens. Then, the vision task used to train the ViT is recovering the tokens of corrupted patches, where each visual token corresponds to an image patch. For BEITv2, it proposes a vector-quantized knowledge distillation based on BEIT to further improve the representation capability of images, which effectively reduces the sensitivity of visual features to changes in image details and presents high-level semantics.

Therefore, the representation learning process of global appearance variation can be expressed as follows.
\begin{equation}
\mathbf{f}_{glob}=F_t([v(f_1),\cdots,v(f_T)])
\end{equation}
where $v(\cdot)$ denotes a pre-trained BEiTv2 model and $F_t(\cdot)$ is a trainable temporal transformer, where extracted visual features are used as input tokens with their temporal positions $t$.

\subsection{Channel Attention Based Feature Fusion}

In Section \ref{sec:local} and \ref{sec:global}, two types of representations are extracted to expose temporal defects caused by the AI video generation process. Due to the diversity of generator architectures and training strategies, AI-generated videos may exhibit various defects. To achieve more reliable detection capability, a channel attention (CA) based fusion module is designed to fuse $\mathbf{f}_{local}$ and $\mathbf{f}_{glob}$, as shown in Fig. \ref{fig:framework}.

%\begin{figure}[]
%\centering
%\includegraphics[width=0.4\textwidth]{fusion.png}
%\caption{The channel attention based feature fusion module.}
%\label{fig:fuse}
%\end{figure}

\begin{equation}
\mathbf{\rho}_a = \delta(\text{BN}(W_2\cdot(\text{ReLU}(\text{BN}(W_1\cdot\mathbf{f}_t)))))
\end{equation}

\begin{equation}
\mathbf{f}_a = \mathbf{\rho}_a\otimes \mathbf{f}_{local}+(\mathbf{1}-\mathbf{\rho}_a)\otimes \mathbf{f}_{glob}
\end{equation}
where $\mathbf{f}_t=\mathbf{f}_{local}\oplus\mathbf{f}_{glob}$. $\oplus$ and $\otimes$ denote the element-wise summation and Hadamard product, respectively. $W_1\in\mathbb{R}^{\frac{C}{r}\times C}$ and $W_2\in\mathbb{R}^{C\times \frac{C}{r}}$ denotes the parameters of fully-connected layers, which are used as channel
context aggregators to exploit channel interactions. $r$ is the channel reduction ratio. $\text{BN}(\cdot)$ and $\text{ReLU}(\cdot)$ denote batch normalization layer and non-linear activation layer, respectively. $\delta(\cdot)$ denotes the Sigmoid function. $\mathbf{1}$ is the all-one vector. In this way, CA-based feature fusion module can adjust the significance of different channels to extract more generalized forensics clues. Then, the cross-entropy loss is considered to optimize the detection network.

\section{Experiments}
In this section, we will evaluate the detection performance of spatial and temporal-based detection algorithms. For AI-generated video forensics in the open world, detectors need to be generalized for fake videos created by unseen generators. Besides, it is also significant to be robust against video lossy operations. Therefore, the generalization and robustness are evaluated in experiments.
%\subsection{Experiment Setting}
%\subsubsection{Generator and Post-processing Cross Test Configuration}
%In order to investigate the generalizability and resilience of the trained detectors, we devised two sets of cross tests. The initial set, known as the cross-generator test, involved training the detector on a given image dataset generated by a particular generator model. Subsequently, we evaluated the performance of the trained detector on independent datasets produced by three distinct generator models, thus enabling an analysis of the detector's ability to generalize amidst varying image generators. The second set of tests referred to as the post-processing cross-test, aimed to emulate real-world implementation scenarios where the examined video samples may undergo damage during transmission across different channels. To achieve this, we employed three different post-processing methods, namely H.265 ABR Compression, H.265 CRF Compression, and Bit Error, each applied with varying degrees of severity to degrade the quality of the test samples. Subsequently, we employed these degraded samples to assess the robustness of the detectors trained on videos of normal quality. Specifically, for each detector trained on a single generator, a total of 16 distinct tests were conducted to capture the full scope of the analysis. During the training process, the detectors utilize temporal information are set to have a resolution of 3 frames, and all detectors are trained with a batch size of 64, a learning rate of $10^{-3}$ and Adam optimizer.

\subsection{Baseline Methods}
%Currently, there is a lack of detection algorithms for AI-generated video. Since identifying AI-generated video can be treated as a binary classification task (real and fake), we first selected several typical video classification models for comparison, including: I3D \cite{carreira2017quo}, S3D \cite{xie2018rethinking} and their modified versions with residual connections (referred to as I3D\_resnet \cite{YiYLTK21} and S3D\_resnet \cite{hara2018can}).

Currently, there is a lack of detection algorithms designed for AI-generated videos. Therefore, we first select two advanced AI-generated image detection methods, including CNNSpot\cite{WangW0OE20} and NPR\cite{2312-10461}. CNNSpot is a typical detection method in the data-driven manner to expose spatial artifacts while NPR is the SOTA method that leverages neighboring pixel relationships to identify both GAN- and diffusion-generated images. Besides, video classification networks, including I3D \cite{carreira2017quo}, S3D \cite{xie2018rethinking}, and their extended versions with residual connections (I3D\_Res and S3D\_Res) are also considered for comparison.

%\begin{enumerate}
%\item[] \textbf{I3D} This network structure was first proposed for action recognition tasks. By expanding convolutional filters and other structures of 2D networks (such as Inception) in the temporal dimension, the inflated network can achieve promising performance for extracting spatial and motion features jointly.
%\item[] \textbf{S3D} Compared with I3D, S3D replaces 3D convolutions with spatial and temporal separable 3D convolutions. By separating space and time, S3D obtain better computational efficiency and performance of video classification tasks.

%\end{enumerate} 

\subsection{Experimental Settings}
In experiments, videos are divided into three groups for training, validation, and testing with the ratio 8:1:1. Videos are decoded into frame sequences, where each video clip is constructed by including seven temporally adjacent frames. 
%Consequently, in the training stage, there are \textcolor{red}{29407} real video clips and \textcolor{blue}{red} fake video clips, where the number of fake video clips generated by each video generator is \textcolor{red}{2563}. 
The pre-trained frame predictor in \cite{HuHHX023} is used to obtain $\mathbf{f}_{local}$. In the CA-based fusion module, $r$ is set as 4. For all detectors, batch size is set as 64. The initial learning rate is set to $10^{-3}$. Adam optimizer is used to update network parameters. 
%Hyperparameters of other methods are as set default. 
The clip-level detection accuracy is used as the evaluation metric, where the threshold of prediction probability is set to 0.5. Please note officially released detectors in CNNSpot\cite{WangW0OE20} and NPR\cite{2312-10461} fail to detect our samples (accuracies are nearly 50\% on average) and they are retrained on our proposed dataset.

\subsection{Performance Evaluation on Cross-generator Scenarios}

In this experiment, the generalization capability of detection methods for unseen video generators is evaluated. In the training phase, fake video clips obtained by a particular video generator are used to train detectors. Then, testing samples produced by all generators are used to evaluate performance.
%In the testing phase, samples produced by the remaining three video generators are used to conduct performance evaluation. 
Experimental results are presented in Table \ref{tab:table1}.

%\subsubsection{Frame Verse Temporal Detection Result}
\begin{table}[h]
\caption{Performance evaluation of generalization capability cross different generators (\%).\label{tab:table1}}
\centering

\begin{tabular}{@{}c|c|cccc|c@{}}
\toprule
\multicolumn{2}{c}{Train\textbackslash{}Test}   & Ali-vilab & Potat1 & ZScope & T2V-zero & Ave. \\ \midrule
\multirow{7}{*}{\begin{tabular}[c]{@{}c@{}}Ali\\ -vilab\end{tabular}}      & CNNSpot     & 97.42& 91.53& \bf{83.68}& 50.18&   80.70\\
                                 & NPR     &   97.48&  85.17&   79.78&   49.81&   78.06\\ \cmidrule(l){2-7}
                                 & I3D          & 96.55& 77.53& 59.85& 62.21&   74.03\\
                                 & I3D\_Res   & 97.90& 93.86& 72.31& 72.98&  84.26\\
                                 & S3D          & 96.80& 86.37& 70.20& 62.29&   78.92\\
                                 & S3D\_Res   & 97.98& 92.43& 73.07& 88.89&   88.09\\
                                 & Ours   &   \bf{99.41}& \bf{94.03}&    77.36&   \bf{93.02}&    \bf{90.96}\\\cmidrule(l){1-7} 
\multirow{7}{*}{Potat1}          & CNNSpot     & 77.60& 98.45& 87.42& 65.18&   82.16\\
                                 & NPR     &   73.34&  97.98&    79.40&    53.14&  75.96\\  \cmidrule(l){2-7}
                                 & I3D          & 70.04& 93.44& 84.52& 71.39&   79.85\\
                                 & I3D\_Res   & 73.74& 95.46& 75.59& 61.62&  76.60\\
                                 & S3D          & 80.22& 98.07& 80.73& 54.97&    78.50\\
                                 & S3D\_Res   & 80.64& 98.82& 76.09& 76.77&   83.08\\ 
                                 & Ours   &   \bf{94.45}&  \bf{99.33}&   \bf{90.24}&  \bf{82.16}&    \bf{91.55}\\\cmidrule(l){1-7} 
\multirow{7}{*}{ZScope}       & CNNSpot     & 67.09& 90.54& \bf{98.15}& 52.26&   77.01\\
                                 & NPR     &   72.20&  85.94&    97.98&   52.39&  77.13\\\cmidrule(l){2-7}
                                 & I3D          & 60.70& 82.25& 95.38& 49.76&    72.02\\
                                 & I3D\_Res   & 78.38& 92.18& 97.40& 52.45&  80.10\\
                                 & S3D          & 67.01& 92.60& 97.65& 51.86&  77.28\\
                                 & S3D\_Res   & 78.13& 90.92& 97.32& 57.76&    81.03\\ 
                                 & Ours   &  \bf{94.71}&  \bf{98.92}&   97.74&    \bf{87.97}&   \bf{94.83}\\\cmidrule(l){1-7} 
\multirow{7}{*}{\begin{tabular}[c]{@{}c@{}}T2V\\ -zero\end{tabular}} & CNNSpot     & 50.00& 50.02& 50.04& \bf{100.00}&  62.52\\
                                 & NPR     &   58.61&  67.15&   54.95&    98.89&  69.90\\\cmidrule(l){2-7}
                                 & I3D          & 50.00& 51.01& 50.34& \bf{100.00}&   62.84\\
                                 & I3D\_Res   & 50.00& 50.00& 50.00& \bf{100.00}&  62.50\\
                                 & S3D          & 50.00& 58.08& 54.38& \bf{100.00}&   65.61\\
                                 & S3D\_Res  & 51.18& 64.14& 60.44& \bf{100.00}&    68.94\\  
                                 & Ours   &   \bf{86.88}&  \bf{95.29}&    \bf{88.56}&    99.33&   \bf{92.52}\\\midrule
\end{tabular}
\end{table}

As shown in Table \ref{tab:table1}, when training and testing samples are generated by the same generator, all methods can obtain a promising detection performance (about 98\% on average). However, for cross-generator cases, spatial domain-based methods (CNNSpot and NPR) and traditional video classification networks suffer from distinct performance drops. In contrast, our proposed detection still achieves satisfactory accuracies in all cases, which demonstrates the efficiency of leveraging both local and global defects to expose AI-generated videos. It should be noted that the cross-generator scenario involving T2V-zero generator is the most challenging case. This result may be due to the fact that T2V-zero is developed based on the pre-trained image diffusion model and then calculates the temporal continuity, which exhibits severe domain shift compared with samples of other generators.

\subsection{Robustness Evaluation Against Video Lossy Operations\label{sec:lossy}}

%In real-world forensic scenarios like social networks, video samples may undergo quality degradation during transmission across different channels. 
In this experiment, the robustness against lossy operations is evaluated.
%To evaluate the robustness of detection methods and simulate lossy operations during network transmission, we employed three different post-processing operations, namely H.265 Average Bitrate Rate (ABR) compression, H.265 Constant Rate Factor (CRF) compression, and Bit Error. Each operation applies various degrees of operation intensity. 
Specifically, for the levels of severity $\{1,2,3\}$, Bit Error considers each of $\{10,5,3\}\times 10^5$ bytes with a bit error; H.265 ABR applies $\{50\%, 25\%, 12.5\%\}$ of the original bit rate; H.265 CRF sets quality factors as $\{27,33,39\}$. Samples from videos of normal quality (without conducting lossy operations) are used to train detectors, and degraded samples are used to assess their robustness. %\textcolor{red}{We consider the challenging case, where detectors are trained on samples generated by text2video-zero and evaluated on other generators' samples.}

\begin{table}[h]
\caption{Robustness evaluation against different lossy operations (\%).\label{tab:table2}}
\centering

\begin{tabular}{@{}c|c|ccccc@{}}
\toprule
\multicolumn{2}{c}{Lossy operations}   & CNNSpot & NPR  & I3D\_Res  & S3D\_Res & Ours\\ \midrule
Raw Data      & None     & 75.60& 75.26& 75.87& 80.29&   \bf{92.46}\\\midrule
\multirow{3}{*}{Bit Error}      & 1     & 73.12& 73.69& 73.49& 79.56&   \bf{89.65}\\
                                 & 2     &   71.80&  72.20&   71.94&   78.81&   \bf{90.31}\\ 
                                 & 3     & 70.34& 70.68& 70.06& 78.04&   \bf{90.07}\\\midrule
\multirow{3}{*}{\begin{tabular}[c]{@{}c@{}}H.265\\ ABR\end{tabular}}& 1   & 75.04& 69.42& 75.78& 79.97& \bf{91.21}\\
                                 & 2          & 74.72& 67.99& 75.46& 79.52&   \bf{89.45}\\
                                 & 3   & 74.11& 67.45& 74.81& 78.58&  \bf{86.71}\\\midrule
\multirow{3}{*}{\begin{tabular}[c]{@{}c@{}}H.265\\ CRF\end{tabular}}& 1   &   74.76& 68.32&    75.53&   79.79&    \bf{90.93}\\
                                & 2   &   74.18& 67.44&   74.80&   78.87&    \bf{88.24}\\ 
                                & 3   &   73.43& 66.75&   73.67&   77.37&    \bf{84.26}\\\cmidrule(l){1-7} 
\end{tabular}
\end{table}

As shown in Table \ref{tab:table2}, all methods obtain their best results for raw data and suffer from a distinct performance degradation against lossy operations. In general, temporal detection methods have better robustness than spatial detection methods since temporal forensics clues, such as global defects, are more likely to survive under lossy operations. Besides, detection accuracies decrease with the increment of severity levels, where distortions caused by lossy operations become more severe on forensic clues. This result infers that it is necessary to develop robustness improvement strategies.

\subsection{Ablation Study}
In this section, key components of the proposed detection framework are evaluated incrementally, including $\mathbf{f}_{glob}$, $\mathbf{f}_{local}$ and the channel attention-based feature fusion (CA-based feature fusion). $\mathbf{f}_{glob}$ is first evaluated independently and then combined with $\mathbf{f}_{local}$ using the simple feature concatenation. Finally, the CA-based fusion is applied to fuse $\mathbf{f}_{glob}$ and $\mathbf{f}_{local}$. 

As shown in Table \ref{tab:ablation}, only using $\mathbf{f}_{glob}$ can expose fake videos to some degree due to the limited generation ability of current video generators. When $\mathbf{f}_{local}$ is considered as the complement, detection performance is improved. Furthermore, the CA-based feature fusion can obtain a distinct performance gain by combining local and global temporal traces adaptively. This result infers that the properties of temporal defects in fake videos generated by different generators are various, which should be taken into consideration for obtaining more reliable detection results.

\begin{table}[]
\centering
\caption{The ablation study of key components (\%).}
\label{tab:ablation}
\begin{tabular}{|c|c|c|c|}
\hline
$\mathbf{f}_{glob}$ & $\mathbf{f}_{local}$ & CA-based fusion & Acc \\ \hline
 \checkmark &   &   & 88.56\\ \hline
 \checkmark  &  \checkmark  &   & 89.52\\ \hline
  \checkmark &  \checkmark  &  \checkmark  & 92.46\\ \hline
\end{tabular}
\end{table}

\section{Conclusion}
To address the security issue of AI-generated videos, we constructed a benchmark video dataset, where samples are generated by several advanced diffusion-based video generators with various contents. Besides, typical video lossy operations are considered to create degraded samples. Then, a novel detection framework is proposed by jointly learning local motion information and global appearance variation, which captures defects at different spatio-temporal scales. To mimic the practical forensics scenarios, extensive experiments are conducted in aspects of generalization and robustness. This work can serve as a baseline for future work about video forensics in the upcoming AIGC era. 

%\begin{thebibliography}{1}
\bibliographystyle{IEEEtran}

\bibliography{reference.bib}

%\bibitem{ref1}
%M. Zhu et al., “Genimage: A million-scale benchmark for detecting AI-generated image,” arXiv.org, https://arxiv.org/abs/2306.08571 (accessed Jan. 2, 2024).

%\bibitem{ref2}
% Zhang, X.; Karaman, S.; Chang, S.-F. Detecting and simulating artifacts in gan fake images. 2019 IEEE international workshop on information forensics and security (WIFS). 2019; pp 1–6.

%\bibitem{ref3}
%Yu, L., Cheng, Y., Sohn, K., Lezama, J., Zhang, H., Chang, H., Hauptmann, A. G., Yang, M.-H., Hao, Y., Essa, I., & Jiang, L. (2023). Magvit: Masked generative video transformer. 2023 IEEE/CVF Conference on Computer Vision and Pattern Recognition (CVPR).

%\bibitem{ref4}
%video diffusion

%\bibitem{ref5}
%text2video-zero

%\end{thebibliography}

\newpage

\vfill

\end{document}